\documentclass[a4paper]{article}
\usepackage{INTERSPEECH2018,amsmath,graphicx,multirow,tabularx, booktabs,array}
\usepackage{color,soul}

\title{Speech recognition for medical conversations}

\name{\begin{tabular}{c}Chung-Cheng Chiu$^{1}$, Anshuman Tripathi$^{1}$\thanks{$^{1}$Equal contribution}, Katherine Chou, Chris Co, Navdeep Jaitly,\\ Diana Jaunzeikare, Anjuli Kannan, Patrick Nguyen, Hasim Sak, Ananth Sankar$^{2}$\thanks{$^{2}$Work conducted while the author was at Google},\\ Justin Tansuwan, Nathan Wan$^{3}$\thanks{$^{3}$Work conducted while the author was at Google}, Yonghui Wu, Xuedong Zhang$^{4}$\thanks{$^{4}$Work conducted while the author was at Google}\end{tabular}}
\address{Google, LinkedIn$^{2}$}
\email{\{chungchengc, anshumant, kic, chrisco, ndjaitly, dianajzk, anjuli, drpng, hasim, jjtswan, yonghui\}@google.com, ansankar@linkedin.com$^{2}$, nwan@freenome.com$^{3}$, sipida@gmail.com$^{4}$}

\begin{document}
%
\maketitle
\begin{abstract}

In this paper we document our experiences with developing speech recognition for \emph{medical transcription} -- a system that automatically transcribes doctor-patient conversations. Towards this goal, we built a system along two different methodological lines -- a Connectionist Temporal Classification (CTC) phoneme based model and a Listen Attend and Spell (LAS) grapheme based model. To train these models we used a corpus of anonymized conversations representing approximately 14,000 hours of speech. Because of noisy transcripts and alignments in the corpus, a  significant amount of effort was invested in data cleaning issues. We describe a two-stage strategy we followed for segmenting the data. The data cleanup and development of a matched language model was essential to the success of the CTC based models. The LAS based models, however were found to be resilient to alignment and transcript noise and did not require the use of language models. CTC models were able to achieve a word error rate of $20.1\%$, and the LAS models were able to achieve $18.3\%$. 
Our analysis shows that both models perform well on important medical utterances and therefore can be practical for transcribing medical conversations.

\end{abstract}
%
\section{Introduction}
\label{sec:intro}
Transcription in the medical space started with stenographers in the early 20th century.  With the proliferation of ASR and NLP technologies around the mid 1990’s, the healthcare system began to adopt single speaker ASR technology to assist with doctor dictations.  More recently, with the widespread adoption of electronic health record (EHR) systems, doctors are now spending $\sim$6 hours of their 11 hour workdays inside the EHR and $\sim$1.5 hours on documentation alone.
With the growing shortage of primary care physicians~\cite{AAMC2017} and higher burnout rates~\cite{Shanafelt2015mayo}, an ASR technology that could accelerate transcription of the clinical visit seemed imminently useful.  It is a foundational technology, that information extraction and summarization technologies can build on top of to help relieve the documentation burden.

Medical conversations between patients (and possibly a caregiver) and providers have several distinguishing characteristics from normal dictations: (1) it involves multiple speakers (the doctor, patient, and occasionally caregiver) with overlapping dialogue at different distances from microphones and varying quality, (2) it covers a range of speech patterns, accents, background noises, and vocabulary from colloquial to complex domain-specific language.
The ASR has to handle long-form content (from 10 to 45 minutes long) that interweaves clinically salient information with casual chatter.
Developing an ASR system for medical conversations is further complicated by the lack of large corpora of clean, curated data to build systems from.  Data gathered by recording actual conversations results in very noisy data because of all the issue arising in real conversations -- disfluencies, simultaneous speech, low signal to noise, to name a few.  Dealing with these factor requires a significant amount of data pre-processing work, without which it is difficult to train traditional speech recognition systems.  Further, some parts of the conversation are more important than others -- casual conversations between the doctor and patient are not as important as the ones describing underlying symptoms, treatments, etc.

More recently, a neural network based speech recognition system has been built for the medical domain using relatively small medical speech data (270 hours) and has been benchmarked against medical transcriptionists~\cite{Edwards2017}.  Speech recognition systems have been evaluated on a clinical question answering task and it has been shown that domain adaptation with a language model improves the accuracy in interpreting spoken clinical questions significantly~\cite{liu2011amia}.  Language model adaptation using crowdsourced input data has been shown to improve the accuracy of a medical speech recognition system~\cite{salloum2017crowdsourced}.  The efficiency and safety of using speech recognition assisted clinical documentation has been compared against use of keyboard and mouse and it was found that sub-optimal speech recognition accuracy has the potential to cause clinical harm~\cite{hodgson2017amia}.

The recent development of end-to-end speech models provides promising alternatives. The Listen Attend and Spell (LAS) model~\cite{chan2015listen} is an end-to-end model that is able to learn a language model as part of the ASR model itself. In this paper, we compare our experiences in building a speech recognition system with LAS and a traditional HMM based models that uses CTC~\cite{Graves06connectionisttemporal} initialization. We find that the LAS model is quite robust to noisy transcripts and does not require a language model. The phoneme based CTC model, on the other hand, only works well when a significant data cleaning effort is undertaken, and a matched language model is developed for the domain.  Our CTC models achieved a WER of $20.1\%$ while the LAS model achieved a WER of $18.3\%$.

We evaluated the performance of both models on capturing important medical phrases.  We selected a subset of conversations, asked a group of professional Medical Scribe to annotate important phrases in the conversation that are useful in writing medical notes.  The CTC model achieves $~92\%$ precision and $~86\%$ recall with bidirectional models and $~88\%$ precision and $~84\%$ recall with unidirectional models, on all the important phrases.  
We analyzed how effective LAS model was at recognizing drug names mentioned in the conversation, that are used in the treatment, and the model achieved $98.2\%$ recall.

\section{System overview}
\label{sec:system}
We explored two mechanisms for building speech recognition models for this task: one used recurrent neural network with connectionist temporal classification (CTC), and the other used end-to-end models with attention (LAS).

\subsection{CTC}

The CTC system consists of an acoustic model trained with CTC loss with context-dependent phoneme outputs, a $n$-gram language model and pronunciation dictionaries. Decoding is done using a finite-state-transducer-based (FST) decoder. We trained both unidirectional (for streaming recognition) and bidirectional CTC models for this task.

\subsection{LAS}

The end-to-end model with attention consists of an encoder, an attention mechanism, and a decoder~\cite{chan2015listen}.  The encoder takes utterances as input, generates a sequence of hidden states, and the decoder uses the attention mechanism to attend to the encoder output at each prediction step, and then generate grapheme output sequentially.  In this framework the encoder plays the role of an acoustic model which processes the speech utterance and transforms them into a high level representation, and then the decoder attends to that output sequence to generate transcripts.  The decoder uses the attention results and the prediction at the previous step to decide the attention at the current step.

\section{Task definition and data processing}
\label{sec:dataproc}
The de-identified data used in this task was acquired from a large dialogue research organization. Audio of conversations was collected by placing a recording device in clinic, with consent of the patient.  The recorded audio was then compressed to MP3 and sent to human annotators for transcription.
The final transcripts contained speaker turns and speaker codes (doctor, patient, care giver, etc.).  In the transcription process, personal data were de-identified by zeroing out the corresponding audio and using a special tag in the transcript. 
Approximate speaker turn timestamps were also provided. We found that these timestamps were often off by order of seconds. For each conversation we also received metadata that included type of interaction, gender of the doctor and a unique identifier for the doctor.

\subsection{Segmentation}
To make training and testing tractable, we segmented the conversations into speaker turn segments. The accuracy of speaker turn boundaries in our training data was found to play a major role in performance of models trained with these segments. We tried the following approaches to get better speaker turn segments.
\subsubsection{Buffered turn alignment}
\label{sec:bufferedturns}
Training a bidirectional model with the original segments gave $40\%$~WER. To fix these speaker turn alignments we added 1~sec audio buffer at the beginning and end of each turn segment and realigned the buffered audio with the turn transcript. We finally created new speaker turn segments by only keeping the audio that aligned with turn transcript. We removed $~10\%$ of tail segments with the worst acoustic model score (normalized by number of time frames). Training and evaluating on these segments gave $27.5\%$~WER with a bidirectional CTC model.

\subsubsection{Two pass alignment}
\label{sec:2passalignment}
To get better aligned data for training and testing, we force-aligned the entire conversation audio with the conversation transcripts using a two-pass forced alignment approach.

\textbf{First pass: confidence islands}.  We align the conversation to the transcript using confidence-islands approach described in~\cite{6707758}.
We recognized the audio using a constrained FST grammar~\emph{G} constructed as follows.

\begin{enumerate}
\item construct a linear FST where each arc represent a word in the transcript.
\item allow $\epsilon$ transitions from start state to all the states in the FST.
\item All states in the FST are final states.
\end{enumerate}
This G only allows paths that are a contiguous sequence of words in the ground truth transcript.
The audio is segmented into 10 sec chunks and recognized using an out-of-domain voicemail acoustic model and the grammar~\emph{G} constructed above. The recognition results are concatenated and the resulting full recognition text is aligned with the actual human transcript. The consecutive sections (at-least 5 words long) in the recognition result that match the actual human transcript (called confidence islands) are assumed to have trustable word timestamps. For each 10 sec segment recognition, we explicitly exclude the first and last words from any confidence islands, since they could have been partially cut at the fixed 10 sec boundary. The timestamps of the words that did not fall in any confidence islands are fixed in a second pass of forced-alignment. We were able to align $\approx80\%$ of the words using just this pass of force-alignment.

\textbf{Second pass: Fix remaining words}.  For any sequence of unaligned consecutive words from the first pass in ground truth transcript, we compute the corresponding audio segment by using the timestamps of confidence islands that precede and follow this sequence in the ground truth. Since we have most of the words already aligned in first pass, these audio segments were not large and we were able to do a fully-constrained force alignment of segment audio against the sequence of unaligned words. Using this second pass method we get timestamps for $\approx98\%$ of words. Timings for rest of the words are interpolated from the neighbouring words which passed forced-alignment.
Once we have the word-level alignments for the full audio, we segmented it into single speaker turns. We found that the turn level alignments obtained by this method were sufficiently accurate for training acoustic models and did not require any further buffered audio approach as described in section~\ref{sec:bufferedturns}.

\subsection{Training and test split}
\label{sec:datasplit}
The data set comprises of about $90,000$ single channel transcribed conversations between doctors and patients during clinical visits. The total amount of data is $\approx14,000$ hours. The conversations represent $151$~types of medical visits that serve different purposes, such as \emph{Wellness Visit}, \emph{Type II Diabetes}, \emph{Rheumatoid Arthritis}, etc. Each conversation is typically between a single doctor and a patient, sometimes also including a nurse, or family member. On average a conversation was $10$~min long with some exceptional conversations as long as $2$~hrs.

We sampled 100 conversations for our test set, ensuring that each conversation had a different doctor, and there was an equal split of male and female doctors. All other conversations containing these $100$~doctors were excluded, leaving 76,000 conversations for training that had no doctor overlap with the test set. $64$~of the test conversations represented $8$~target disease areas. Disease areas, such as \emph{dermatology} comprise interaction types, such as \emph{eczema}, \emph{melanoma}, and \emph{acne}. 36 conversations were sampled from the non-target disease areas.

For training and testing we used speaker turn segments obtained from the two pass force-alignment explained in section~\ref{sec:2passalignment}. As a general preprocessing, we removed any speaker turn segment that contained de-identified tag in transcript, from both test and training set. Manual inspection showed that most of the issues with segmentation were with very small utterances. To reduce this type of errors we removed any speaker turn smaller than 5 words from our test set. For training data we removed single word turns.

\section{Model training}
\label{sec:modeltraining}
For training both attention models and CTC phoneme based models, the features used is Fast Fourier Transform (FFT) with 80~filterbank channels computed on $25$~ms audio frames with a stride of $10$~ms. As input to the model we stack 3 such frames and take a stride of 3, \emph{i.e.}, input is 240 filterbanks every $30$~ms. For computing the FFTs we 
ignore frequencies above $3800$Hz.

To make the models more robust to noise and prevent over-training we added artificial noise to audio while training, called Multi-style TRaining (MTR)~\cite{1169544}. In our setup each training utterance is combined with 20 different noises (room reverberations, background music, cafe noises) at a SNR ranging from 5dB to 25dB. We found that best results are obtained when noise is added when using CTC/Cross Entropy (CE) training criteria and original audio is used while doing EMBR training~\cite{DBLP:journals/corr/Shannon17}.

\subsection{CTC}

For this task, we trained CTC models with context-dependent (CD) phoneme outputs. The model architecture is a stack of LSTM layers feeding a softmax output predicting 8K CD phone units plus 1 CTC blank symbol. The CD phones were computed as described in~\cite{Young94tree-basedstate} from training data initially based on alignments from an out-of-domain voice-mail model. Once we trained an in-domain model with CE loss on this CD phoneset, we realigned the training data with the CE model and computed a new CD phoneset that was used for training the CTC models. We trained both bidirectional and unidirectional models for this task. Since this is conversational data, segments generally do not end with enough silence to give model the time to output phonemes for all the frames. We found that to get a good unidirectional model we had to train models with output delay, where we ignore the output from the model for first $T$ frames and then repeat the last input frame $T$ times.

During decoding, we used a 5-gram Language model trained on a mixture of medical and voice search/dictation data, using bayesian interpolation~\cite{37567} for combining data from different domains. For pronunciations we used supervised pronunciation dictionaries ($77\%$ of pronunciations) and grapheme-to-phoneme (G2P) models ($23\%$ of pronunciations).

\subsection{LAS}
Our sequence-to-sequence attention model consists of an encoder, an attention mechanism, and a decoder.  The encoder architecture is stacked bi-directional LSTMs.  The bi-directional LSTMs combine the information of both forward and backward directions, leading the output hidden states to have access to the whole utterance, with information peaked at the local utterance frames.  Having access to the whole utterance helps the encoder, as an acoustic model, to distinguish utterance signal from the environmental noise.  The other benefit is to make the attention mechanism easier to learn.  Since each output hidden state contain mostly the local utterance information, with access to the whole utterance for distinguishing noise, the attention mechanism can learn to focus mainly on the frames corresponding to the current prediction target.

Our model use multi-headed attention~\cite{aaswani-nips-2017}, which extends the conventional attention mechanism to have multiple heads, where each head can generate different attention distributions.  We use scheduled sampling~\cite{bengio-arxiv-2015} for training the decoder.  At the beginning we use the ground truth as the previous prediction and as training proceeds we linearly ramp up the probability of sampling from model's prediction up to $30\%$, and then stay at this sampling rate till the end.  The decoder performs grapheme predictions.



\section{Results and Analysis}
\label{sec:results}
We trained LAS attention models and CTC phoneme based models using speaker turn based training data defined in section~\ref{sec:datasplit}. We randomly sampled $\sim 0.5\%$ of utterances from the training data and made that our dev set. CTC and attention models shared the same train/test/dev data.

\subsection{CTC results}

We trained both unidirectional and bidirectional CTC models.  For the unidirectional model we add an output delay of 10 steps. The model has 5 LSTM layers with 700 units per layer, and reached a WER of $28.8\%$ after CTC training criteria and improved to $23.5\%$ with EMBR training. Training a model with 5 bidirectional LSTM layers and 700 units per layer (350 units for either direction) gives $25.2\%$ WER that improves to $20.1\%$ after EMBR training.  Table~\ref{table:ctcresults} shows WER numbers for CTC model on different subsets of the test set.

\subsection{LAS results}

\begin{table}[t]
\caption{WER for different data alignment approaches.  The LAS models here are the basic version without the techniques we applied for the final model.}
\centering
\begin{tabular}{lrr}
\hline
Alignment & CTC Bidi & LAS Bidi \\
\hline
Original & $40$ & - \\
\hline
Buffered & $27.5$ & $23.4$ \\
\hline
Confidence islands & $20.1$ & $22.4$ \\
\hline
\end{tabular}
\label{table:overall}
\end{table}

We tried adding convolution layers before LSTM layers in the encoder when training without MTR.  The resulting model gives $22.4\%$ WER.  Adding scheduled sampling reduce it to $21.6\%$.  Latter when adding MTR, the convolution layers did not help and tend to provide worse quality, and therefore we remove the convolution layers from the encoder.  With MTR and without convolution layers, the WER goes to $18.9\%$.  Adding multi-headed attention leads to $18.3\%$ WER.  We also tried EMBR training but it did not improve WER further.  The encoder has $6$ layers of bi-directional LSTMs with $768$ units for each direction, and the decoder has $3$ layers of uni-directional LSTMs with $768$ units.

\subsection{Comparing CTC and LAS}

Attention models and CTC models of comparable sizes give very similar WER before EMBR training on CTC models. A bigger attention model outperforms CTC model. Although both CTC and attention models perform badly on small utterances, attention models performs better than CTC. We found that on very small utterances like ``yeah'' attention model gets the output correct most of the time even though the audio is incomprehensible. The attention model also performs better when audio is truncated abruptly at beginning or end of the utterance.  In general attention model is more resilient to the data processing.

\subsection{Error analysis}
In isolation utterances are often completely incomprehensible, transcribers likely used context to understand them.  Lots of short words are filled in by transcribers whether or not they are audible, {\em e.g.}~``had blood work'' is transcribed as ``I had blood work''. Vocal qualities make language hard to understand, {\em e.g.}~accents, laughing, whispering, etc.  Patient speech is often softer and more distant, mostly due to the recording setup). The audio was also compressed to MP3 which likely caused loss in quality.
\subsubsection{CTC}

\newcolumntype{L}{>{\centering\arraybackslash}m{2cm}}
\begin{table}[t]
\caption{Analyzing CTC results}
\centering
\begin{tabular}{Llrrr}
\hline
\multirow{2}{*}{Category} & \multirow{2}{*}{Segments} & \multirow{2}{*}{Words} & \multicolumn{2}{c}{WER} \\
\cline{4-5}
&  &  & Unidi & Bidi \\
\hline
\multirow{1}{*}{All} & All & 134,878 & 23.5 & 20.1 \\
\hline
\multirow{2}{*}{Speaker} & Doctor & 77,851 & 20.6 & 17.4 \\
                         & Patient & 53,740 & 27.0 & 23.3\\
\hline
\multirow{2}{*}{Gender} & Male & 64,615 & 23.1 & 19.8\\
                        & Female & 70,263 & 23.9 & 20.4\\
\hline
\multirow{8}{*}{\parbox{2cm}{\centering Target disease area}} & Cardiology & 12,066 & 24.3  & 20.6\\
                              & Dermatology & 9,657 & 21.1  & 18.2\\
                              & Diabetes & 9,569 & 20.8 & 17.8\\
                              & Mental health & 11,636 & 21.8 & 19.0\\
                              & Oncology & 9,817 & 27.1 & 25.4\\
                              & Primary care & 9,635 & 24.9 & 24.3\\
                              & Pulmonology & 11,742 & 25.2 & 21.0\\
                              & Urology & 9,061 & 21.1 & 18.7\\
\hline
\end{tabular}
\label{table:ctcresults}
\end{table}

Most of the errors seem to appear at the beginning and end of the utterances, {\em e.g.}: ``that's all you needed'' becomes ``that's all you need it''. Anecdotally it was observed that when the speakers are talking over each other, the CTC model deletes words in that region of audio. A major portion of errors is accounted by small speaker turn segments ($<1$sec long), for longer utterances model misses out on conversational part of the speech {\em e.g.}: ``uh i am'' transcribed as ``um i mean''. In general the WER on patient speaker turns is worse than doctors, this is possibly because the recording device was placed closer to the doctors and hence doctors' speech is much clearer than the patient.  A detailed analysis with respect to speaker, gender, and disease area is shown in Table~\ref{table:ctcresults}.

To see the performance of the model on important medical phrases, we collected a supervised list of medical phrases from ground truth transcripts of these conversations and did evaluation of how often do we recognize those phrases correctly in segments of audio. For collecting these phrases, a group of Medical Scribe were asked to mark important phrases in the conversation that are useful in writing medical notes ({\em e.g.}: symptoms, lab results, patient directions, etc). The results showed that CTC model based recognition achieves $~92\%$ precision and $~86\%$ recall with bidirectional models and $~88\%$ precision and $~84\%$ recall with unidirectional models on recognizing important medical phrases. It is also noteworthy that some information, particularly patient instructions,
are oftentimes repeated by the doctor, to ensure comprehension by the patient, further mitigating errors.

\subsubsection{LAS}

On scanning through the transcript errors, most of them are conversational and unrelated to medical terms.  Among the errors related to medical terms, it is usually more related to acoustic modeling than lack of medical terms.  For example, transcribe ``Don't pay a penny'' as ``Don't byetta penny'', where the model replaces a common word with a medical term.  LAS does not use an external language model, and does have less casual conversational content to learn from.

We analyzed how effective LAS model is at recognizing drug names mentioned in the conversation, that are used in the treatment, and the model achieved $98.2\%$ recall.

\begin{table}[t]
\caption{Modeling results.  All CTC results are with MTR, where the base LAS model is without.  All LAS results are without EMBR training.  *We re-tuned LAS's architecture when adding MTR, and therefore the improvement of the error rate reflect both changes.}
\centering
\begin{tabular}{lrrr}
\hline
\multirow{4}{*}{CTC}              & \multirow{2}{*}{Unidi} & Base (MTR) & $28.8$ \\
                                  &                        & $+$ EMBR & $23.5$ \\
                                  & \multirow{2}{*}{Bidi}  & Base (MTR) & $25.2$ \\
                                  &                        & $+$ EMBR & $20.1$ \\
\hline
\multirow{4}{*}{LAS}              & \multirow{4}{*}{Bidi}  & Base & $22.4$ \\
                                  &                        & $+$ MTR* & $19.7$ \\
                                  &                        & $+$ Scheduled sampling & $18.9$ \\
                                  &                        & $+$ Multi-headed attention & $18.3$ \\
\hline
\end{tabular}
\label{table:modeling}
\end{table}

\section{Conclusions}

In this work we explored building automatic speech recognition models for transcribing doctor patient conversation.  We collected a large scale dataset of clinical conversations ($14,000$ hr), designed the task to represent the real word scenario, and explored several alignment approaches to iteratively improve data quality.  We explored both CTC and LAS systems for building speech recognition models.  The LAS was more resilient to noisy data and CTC required more data clean up.
A detailed analysis is provided for understanding the performance for clinical tasks.  Our analysis showed the speech recognition models performed well on important medical utterances, while errors occurred in causal conversations.  Overall we believe the resulting models can provide reasonable quality in practice.

\section{Acknowledgement}

We would like to thank Jack Po, Michael Pearson, and Kasumi Widner for data business development and management, and Francoise Beaufays, Claire Cui and Jeff Dean for high-level sponsorship of this research.

\bibliographystyle{IEEEtran}

\bibliography{main}

\end{document}